\documentclass[conference]{IEEEtran}
\IEEEoverridecommandlockouts
\usepackage{cite}
\usepackage{amsmath,amssymb,amsfonts}
\usepackage{graphicx}
\usepackage{textcomp}
\usepackage{xcolor}
\usepackage{soul}
\usepackage{subcaption}
\usepackage{footnote}
\usepackage{hyperref}
\usepackage{csquotes}
\usepackage{algorithm}
\usepackage{algpseudocode}
\usepackage{multirow}
\usepackage{multicol}

\def\BibTeX{{\rm B\kern-.05em{\sc i\kern-.025em b}\kern-.08em
T\kern-.1667em\lower.7ex\hbox{E}\kern-.125emX}}

\hypersetup{ 
    colorlinks=true,
    linkcolor=black, 
    citecolor=green, 
    urlcolor=blue
}

\begin{document}

\title{NAVIGATING LIMITATIONS WITH PRECISION: A FINE-GRAINED ENSEMBLE APPROACH TO WRIST PATHOLOGY RECOGNITION ON A LIMITED X-RAY DATASET}

\author{
    \IEEEauthorblockN{Ammar Ahmed\IEEEauthorrefmark{1},
    Ali Shariq Imran\IEEEauthorrefmark{1},
    Mohib Ullah\IEEEauthorrefmark{1},
    Zenun Kastrati\IEEEauthorrefmark{2},
    Sher Muhammad Daudpota\IEEEauthorrefmark{3}}
    \IEEEauthorblockA{
    \begin{tabular}{c}
        \IEEEauthorrefmark{1}Intelligent Systems and Analytics (ISA) Research Group, Department of Computer Science (IDI), \\Norwegian University of Science \& Technology (NTNU), Gjøvik, 2815, Norway \\
        \IEEEauthorrefmark{2}Department of Informatics, Linnaeus University, Växjö, 351 95, Sweden \\
        \IEEEauthorrefmark{3}Department of Computer Science, Sukkur IBA University, Sukkur, 65200, Pakistan \\
    \end{tabular}
    }
}

\maketitle

\begin{abstract}
The exploration of automated wrist fracture recognition has gained considerable research attention in recent years. In practical medical scenarios, physicians and surgeons may lack the specialized expertise required for accurate X-ray interpretation, highlighting the need for machine vision to enhance diagnostic accuracy. However, conventional recognition techniques face challenges in discerning subtle differences in X-rays when classifying wrist pathologies, as many of these pathologies, such as fractures, can be small and hard to distinguish. This study tackles wrist pathology recognition as a fine-grained visual recognition (FGVR) problem, utilizing a limited, custom-curated dataset that mirrors real-world medical constraints, relying solely on image-level annotations. We introduce a specialized FGVR-based ensemble approach to identify discriminative regions within X-rays. We employ an Explainable AI (XAI) technique called Grad-CAM to pinpoint these regions. Our ensemble approach outperformed many conventional SOTA and FGVR techniques, underscoring the effectiveness of our strategy in enhancing accuracy in wrist pathology recognition.
\end{abstract}


\section{\textbf{Introduction}}
Hospitals, particularly their emergency services, frequently handle a substantial volume of wrist fracture cases, with wrist pathologies, especially fractures, being a prevalent occurrence in children and adolescents \cite{hedstrom_2010}. While digital radiography is employed to capture X-rays, studies have indicated that diagnostic errors in interpreting emergency X-rays can escalate up to 26\% \cite{Mounts_2011, Er_2013}. These errors stem from various human and environmental factors, including clinician inexperience, fatigue, distractions, suboptimal viewing conditions, and time constraints. The prospect of automated radiograph analysis by computers, characterized by consistency and tirelessness, holds significant promise in complementing the efforts of emergency physicians and radiologists. 

Recently, convolutional neural networks have shown promise in automating pathology recognition in trauma X-rays \cite{adams2020ai, ahmed2024enhancing}. However, distinguishing subtle variations among wrist pathologies in X-rays presents a notable challenge. One potential strategy involves manual annotation to highlight the distinctive regions that showcase these pathologies. However, this method is labor-intensive, expensive, and requires specialized domain knowledge. The crucial question arises: how can we differentiate between visually similar categories without relying on manual annotation and the extensive data typically required for such intricate recognition tasks?

In computer vision, a common approach when dealing with limited data is known as \enquote{hand-engineering}. This encompasses tasks ranging from meticulous manual annotation to the creation of specialized components within CNN architectures \cite{Dhami2018}. Such components can be designed to be optimal for specific recognition challenges, such as ensuring the architecture deliberately seeks the most discriminative regions and identifies patterns even when dealing with minimal data.

In this study, we hypothesize that the base FGVR architecture proposed by Chou et al. \cite{Chou2022} will outperform many of the existing conventional and other FGVR recognition techniques on this task of wrist pathology recognition. We then build upon the base method and introduce two variations of the base method, and propose an ensemble approach incorporating all three variations, demonstrating superior performance compared to the base method.
The fundamental design principle is to consider each pixel on a feature map as a distinct feature, as many of the wrist pathologies, such as fractures, can be very small in size. We determine the significance of each pixel based on its predicted class score and subsequently merge the critical pixels. We validate our hypothesis through rigorous testing on the custom-curated GRAZPEDWRI dataset, specifically designed for wrist trauma X-rays \cite{nagy2022pediatric}. To the best of our knowledge, this is the first time a fine-grained analysis has been done on real-world X-rays involving different projections and the presence of several pathologies.

\section{\textbf{Related Work}}
Although numerous studies have been conducted on wrist fracture detection, limited attention has been given to wrist pathology classification. We explored related works concerning fracture detection and classification over the past three years. Guan et al. \cite{Guan2020} employed a two-stage R-CNN, achieving an AP of 0.62 across approximately 4,000 arm fracture X-ray images from the MURA dataset. Kandel et al. \cite{kandel2020musculoskeletal} fine-tuned six CNNs for musculoskeletal image classification; their fine-tuned DenseNet121 achieved a mean accuracy of 0.82. Wang et al. \cite{wang_yao_zhang_guan_wang_2021} introduced a two-stage R-CNN structure with a TripleNet backbone to detect fractures. They tested this on a collection of 3,842 thigh fracture X-ray images and achieved an AP of 0.88. Xue et al. \cite{Xue2021} introduced a guided anchoring approach with Faster R-CNN. Their evaluation, based on 3067 images, demonstrated an AP of 0.71. Raisuddin et al. \cite{raisuddin2021} introduced DeepWrist for distal radius fracture detection, achieving AP scores of 0.99 and 0.64 on two test sets of 207 and 105 challenging cases. Despite generating heatmaps indicating fracture likelihood, the model faced limitations in precise fracture localization due to a small and imbalanced dataset. Ma et al. \cite{ma2021bone} initially classified Radiopaedia dataset images into fracture and non-fracture using CrackNet. Subsequently, they employed Faster R-CNN for 1052 bone images, achieving an accuracy of 0.88. Kim et al. \cite{kim2021} collected wrist trauma images from the emergency department, utilizing DenseNet-161 and ResNet-152. DenseNet-161 achieved the highest sensitivity and accuracy of 0.90, with an AU-ROC of 0.96 for wrist fracture detection. Joshi et al. \cite{Joshi2022} employed transfer learning with a modified Mask R-CNN on 3000 surface crack images and 315 wrist fracture images, achieving 0.92 AP for detection and 0.78 for segmentation. Hardalac et al. \cite{hardalac2022} conducted 20 fracture detection experiments using wrist X-ray images, achieving the highest AP of 0.86 with their ensemble model, WFD-C. Hržić et al. \cite{hrzic2022fracture} compared YOLOv4 and U-Net for fracture detection on the GRAZPEDWRI-DX dataset, with YOLOv4 outperforming with higher AUC-ROC (up to 0.90) and F1 scores (up to 0.96).

It can be seen from the recent literature that most studies employ object detection relying on a small, manually annotated dataset. In our work, we take a fine-grained recognition approach to identify wrist pathologies. We highlight discriminative X-ray regions using XAI and eliminate the need for manual bounding boxes. 

\section{\textbf{Material \& Methods}}
\subsection{Dataset Curation}
\begin{figure}
\includegraphics[width=1\linewidth, height=5.5cm]{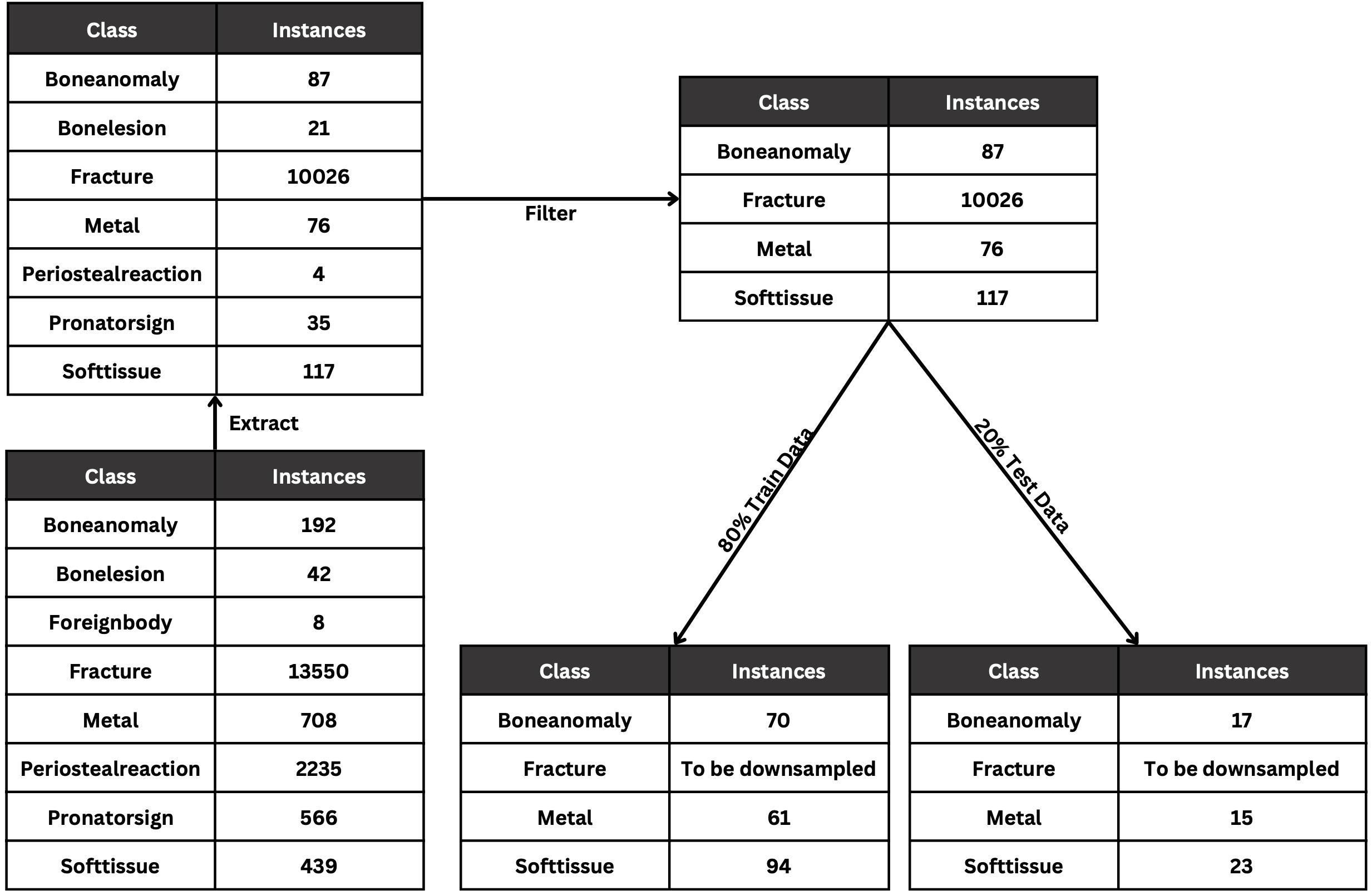}
\caption{Illustration of dataset curation steps.}
\label{fig1:dataset_curation}
\end{figure}

We customized a dataset derived from GRAZPEDWRI, featuring 20,327 wrist images from 6,091 patients aged 0.2 to 19 years. The dataset includes both lateral and posteroanterior projections, posing challenges such as instances with multiple objects and significant class imbalance, especially in the \enquote{fracture} class. Our focus on multi-class recognition led us to extract single-class images to address this challenge. As shown in Fig.\ref{fig1:dataset_curation}, we address the former issue by selectively extracting images representing single classes, excluding the \enquote{foreignbody} class due to limited instances, significantly reducing the total instances. Next, filtering classes with few instances results in four final classes. For training and testing, 20\% of data from each class (except \enquote{fracture}) is allocated for testing. To address the class imbalance, we employ a downsampling strategy for the \enquote{fracture} class, and the extent of this downsampling is contingent on the number of augmentations applied to each class. For testing purposes, we adopt two distinct approaches: in one testing set, we augment the existing images to reach a count of around 120 images per class, while in the other, we keep the original images for each class (deemed as challenging test set) and reduce the \enquote{fracture} class to 120 and 25 images respectively. The first test set introduces variations and perturbations to the original data. Testing on this set evaluates how well the model generalizes to diverse and transformed instances. For the training set, we augmented each class to around 500 images, this struck a balance between data quality and diversity. Table \ref{tab:testsets} shows the number of instances for each class in all sets after the entire curation process.
\begin{table}[h]
\centering
\caption{Number of instances in the subsets.}
    \begin{tabular}{l c c c}
    \hline
    Class & Training Set (After Aug) & Test Set 1 & Test Set 2 \\
    \hline 
  Boneanomaly  & 490 & 119 & 17 \\
  Fracture  & 500 & 120 & 25 \\
  Metal  & 496 & 120 & 15 \\
  Softtissue  & 470 & 115 & 23 \\
    \hline     
    Total & 1956  & 474 & 80\\
    \hline
  \end{tabular}
  \label{tab:testsets}
\end{table}
\begin{figure*}
\centering 
\includegraphics[width=0.7\textwidth, height=5cm]{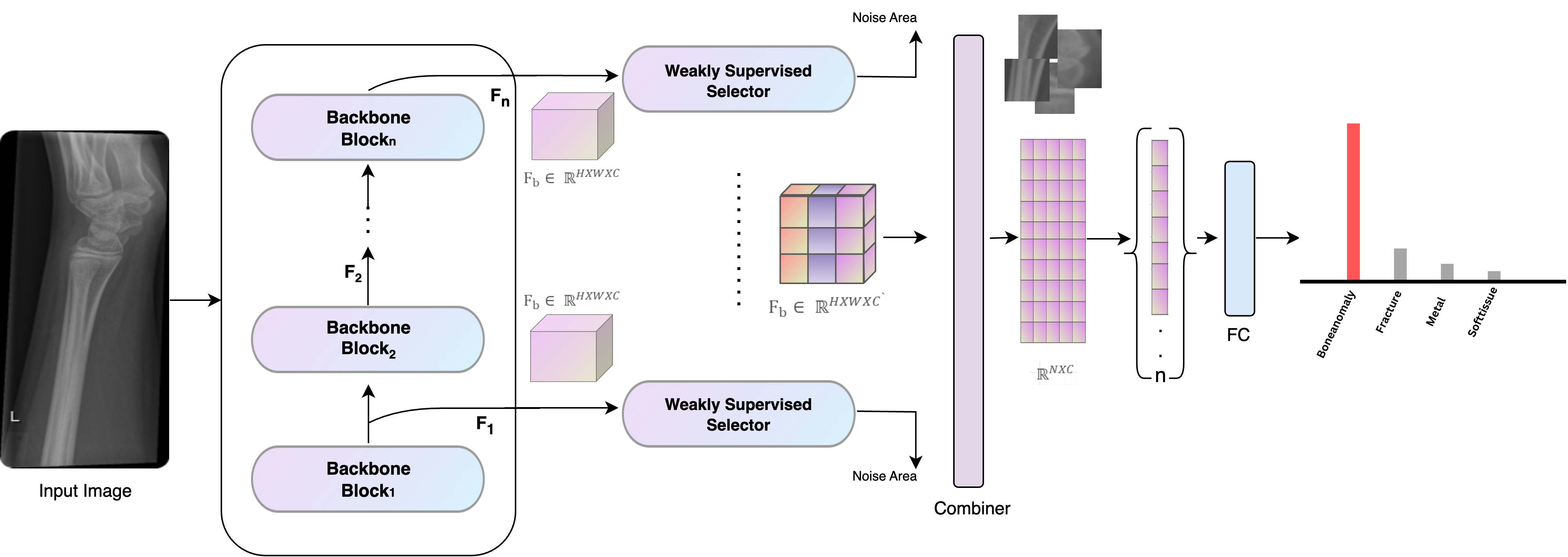}
\caption{Pipeline of plug-in module for fine-grained wrist pathology recognition.}
\label{fig1:pim_architecture}
\end{figure*}
Out of 1956 training instances, 20\% was reserved as validation data. For the augmentation process, we employed Keras' ImageDataGenerator, utilizing rotation, width shift, height shift, zoom range, horizontal flip, and brightness range. We opted for optimal specific values to strike a balance between augmenting the images effectively without introducing excessive rotation, shift, brightness exposure, or zoom that could potentially lead to images being out of frame or overly exposed. 

\subsection{Plug-in Module For FGVR}
\begin{figure*}
\centering 
\includegraphics[width=0.8\textwidth, height=6cm]{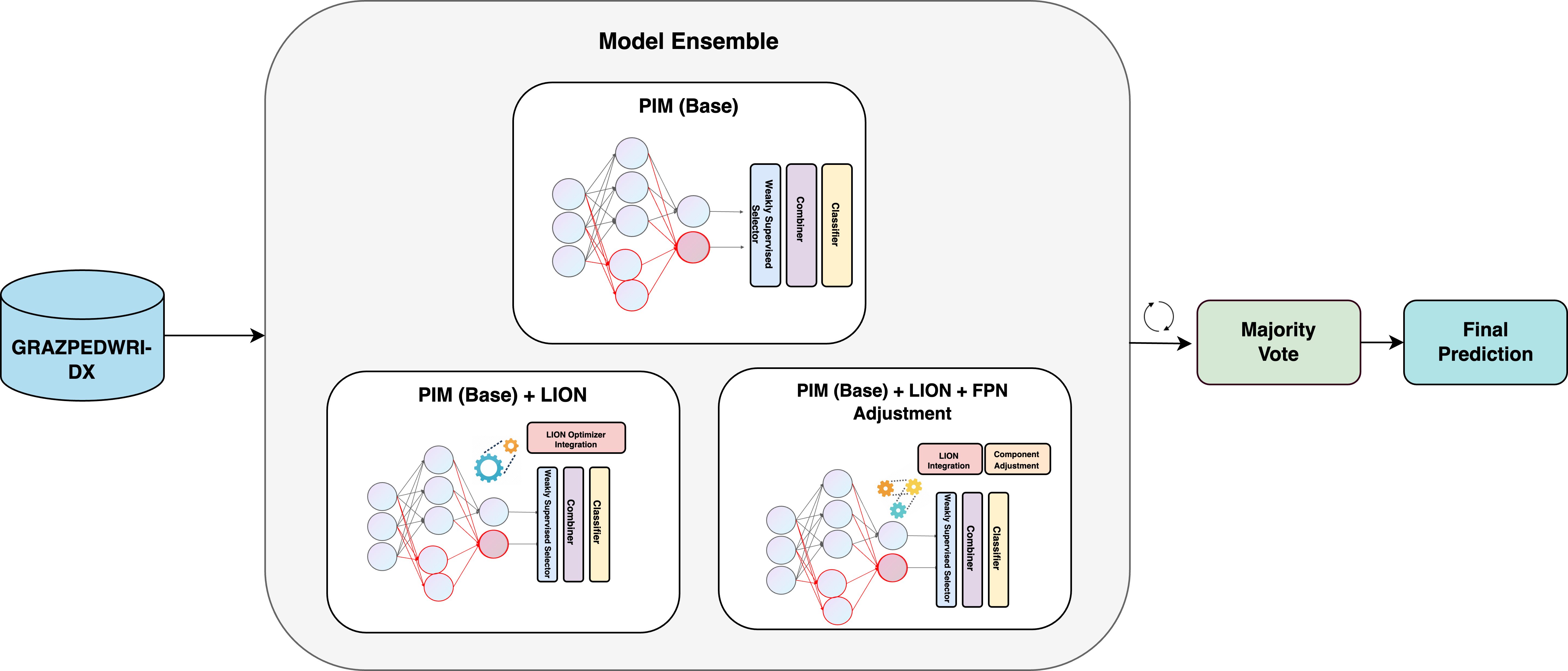}
\caption{Ensemble pipeline for the Plug-in module, incorporating three configurations: base, base + LION, and base + LION + FPN. Final predictions are determined through majority voting.}
\label{fig1:ed}
\end{figure*}

In this study, we demonstrate the potential of fine-grained recognition in the domain of wrist pathology recognition using a framework called Plug-in Module (PIM) for FGVR. The network incorporates innovative background segmentation and feature combination methods, enhancing FGVR precision. We enhance the network's performance further by integrating the \textit{Evo\textbf{l}ved S\textbf{i}gn M\textbf{o}me\textbf{n}tum} (LION) optimizer \cite{Chen2023}, creating a more robust variant. Additionally, we conduct an ablation analysis, exploring various network elements, leading to a second variant with adjusted FPN size and LION integration. The final step involves ensemble learning, combining all three variants (Base + LION + FPN) into a robust network fine-tuned for wrist pathology recognition based on the majority voting ensemble technique. The PIM architecture is shown in Fig. \ref{fig1:pim_architecture}, whereas our ensemble approach pipeline in Fig. \ref{fig1:ed}. 

The Plug-in Module consists of a backbone, selector, combiner, and Feature Pyramid Network (FPN). Recognizing ViT's limitation in hierarchical feature expression for local regions, we opt for SwinTransformer \cite{liu2021swin} as our backbone. Swin-T, chosen for its multi-layer self-attention structure, comprises four blocks, each generating regions selected by the Weakly Supervised Selector.

The Weakly Supervised Selector receives feature maps as input, employing a linear classifier to classify each pixel. This process identifies the most discriminative regions, relegating background noise to a flat probability. The selection criterion is based on the maximum predicted probability, distinguishing pivotal feature points from less relevant ones. This emphasis on the highest likelihood of accurate classification minimizes distraction from background noise. The importance of each pixel \( f_i \) in a particular feature map \( F_i \) can be determined using the softmax function as follows:
\begin{equation}
s(f_i) = e^{f_i} \left( \sum_{j} e^{f_j} \right)^{-1}, \quad \forall f_i \in F_i
\end{equation}
With the probabilities of each feature point \( f_i \), our objective is to select those with the highest probabilities. Denoting the probability of feature point \( f_i \) as \( p_i \) and assuming a total of \( n \) feature points, we initially sort the probability indices in descending order, yielding a sorted index vector \( I \) as follows:
\begin{equation} \label{eq:sort}
I = \text{argsort}^{\downarrow}(\mathbf{p}),
\end{equation}
where \( \mathbf{p} = [p_1, p_2, \ldots, p_n] \) is the vector of probabilities. Next, we select the top \( k \) indices from \( I \) to obtain a subset of indices \( K \) as follows:
\begin{equation} \label{eq:subset}
K = I[1:k].
\end{equation}
Finally, we gather the most important feature points \( f_i \) corresponding to the indices in \( K \) to obtain a set of important feature points \( \mathbf{f}_{\text{important}} \) as follows:
\begin{equation} \label{eq:gather}
\mathbf{f}_{\text{important}} = \{ f_i : i \in K \}.
\end{equation}

To effectively detect objects of varying sizes, especially in X-ray images of wrist pathologies, feature extraction at multiple scales is crucial. Fractures and wrist pathologies can exhibit different sizes, making the use of Feature Pyramid Networks (FPNs) advantageous. FPNs allow the extraction of features at various levels of detail, enabling a comprehensive analysis of objects at different scales. For feature fusion, the method employed is graph convolution, where selected feature points are organized into a graph structure. This graph, representing features at different spatial locations and scales, is input into a Graph Convolutional Network (GCN) to learn relationships among nodes. Subsequently, the feature points are consolidated into super nodes via a pooling layer, their features are averaged, and a linear classifier is used for prediction.

Having outlined the components of the plug-in module, we now explain its operation using an example, referencing Fig. \ref{fig1:pim_architecture}. Let the feature map from the $b^{th}$ block in the backbone network be $F_b \in \mathbb{R}^{C \times H \times W}$, where $H$, $W$, and $C$ are the height, width, and size of the feature dimension. This map is input to a weakly supervised selector, resulting in a new feature map $F_b' \in \mathbb{R}^{C' \times H \times W}$, where $C'$ is the number of target classes. Each feature point is classified, and the points with the highest confidence scores are selected. The selected features are fused through the fusion model. Assuming $N$ selected feature points, the feature maps are concatenated along the feature dimension before input to the fully connected layer:
\begin{equation}\label{eq_concat}
    F_{\text{concat}} = \text{Concat}(f_1', f_2', \ldots, f_N') \in \mathbb{R}^{N \times C'}.
\end{equation}
Subsequently, the concatenated feature map $F_{\text{concat}}$ is input to the fully connected layer, yielding a prediction result of dimension $\mathbb{R}^{C'}$:
\begin{equation}
    F_{\text{pred}} = F_{\text{concat}} W + b \in \mathbb{R}^{C'},
\end{equation}
where $W$ and $b$ denote the weight matrix and bias vector. This architecture combines local features with global features, representing the entire image. Note that for simplicity, Eq. \ref{eq_concat} uses straightforward concatenation, but we use graph convolution for fusion, facilitating efficient integration without compromising results from the backbone model.

\subsection{Evolved Sign Momentum (LION)}
The primary goal of deep neural networks is to minimize the differences between predicted and true values measured by a loss function. Optimizers aim to reduce this loss, improving prediction accuracy. Well-known optimizers include Adam and SGD, which are known for their effectiveness in various contexts. However, Adam's high memory consumption can be problematic for large models or batch sizes. SGD's convergence can be slower, and it is sensitive to feature scaling. To overcome these issues, we utilize the advanced optimizer LION \cite{Chen2023}. LION maintains superior memory efficiency by only tracking momentum and employs a consistent magnitude for each parameter, unlike adaptive optimizers.

\begin{figure}
\includegraphics[width=1\linewidth, height=5.5cm]{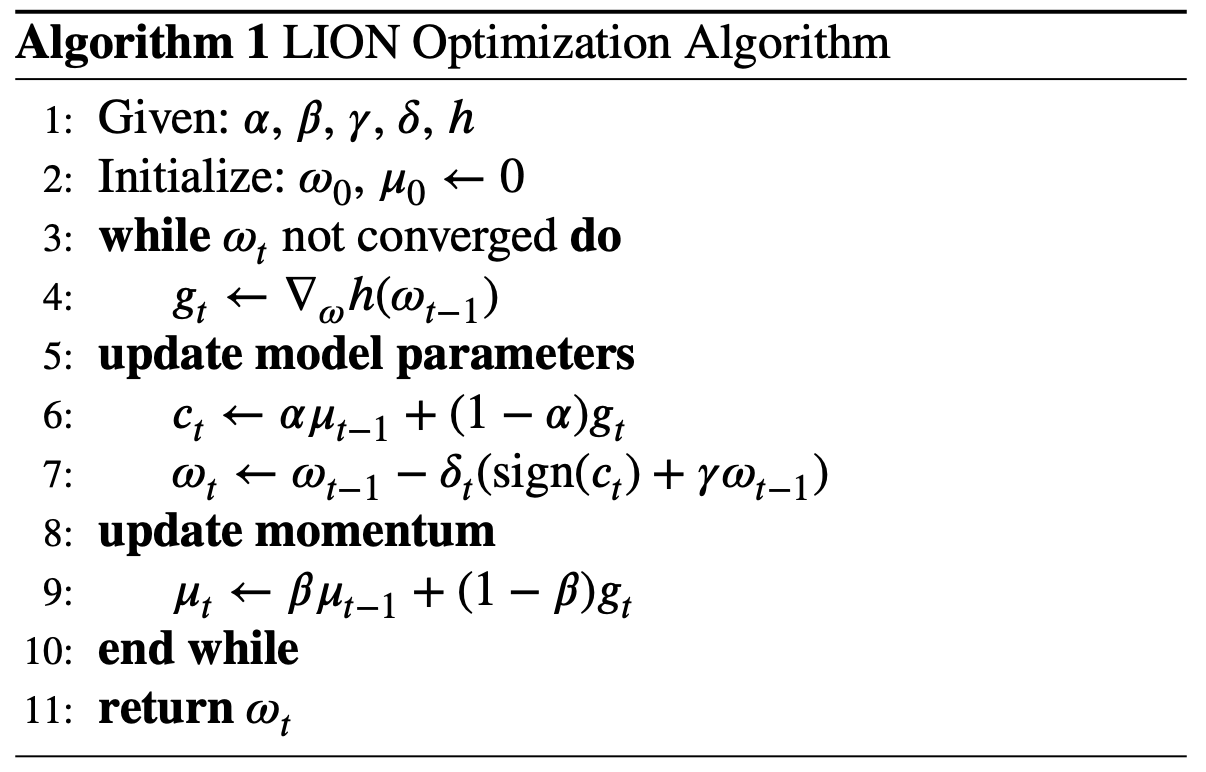}
\caption{LION Optimization Pseudocode.}
\label{fig1:algo}
\end{figure}

In our study, we demonstrate that incorporating LION into the fine-grained recognition plug-in module improves performance compared to using SGD as the default optimizer. Fig. \ref{fig1:algo} provides a pseudocode for LION. Given learning rates $\alpha$, $\beta$, regularization factor $\gamma$, update rate $\delta$, and objective function $h$, the algorithm iterates until convergence, calculating the gradient of $h$ with respect to $\omega$ at $\omega_{t-1}$ at each step. It computes $c_t$ (a mix of the previous momentum $\mu_{t-1}$ and the current gradient $g_t$ for smoothing updates) and updates $\omega$ using a rule with learning rate $\delta_t$, the sign of $c_t$, and a regularization term $\gamma \omega_{t-1}$. The sign function ensures a uniform update magnitude and the regularization term controls parameter values. Momentum $\mu$ is updated for the next iteration, carrying forward information from previous gradients. Once parameters converge, the optimized parameters $\omega_t$ are returned as the output.

\subsection{Experimental Settings}
We complemented our fine-grained approach with a set of established models, each with distinct architectural principles and widely benchmarked in many studies. The models encompass EfficientNetV2 \cite{tan2021efficientnetv2}, EfficientNetb0 \cite{tan2020efficientnet}, VGG16 \cite{simonyan2015verydeep}, ViT \cite{Dosovitskiy2020}, RegNet \cite{xu2021regnet}, DenseNet201 \& DenseNet121 \cite{huang2018densely}, MobileNetV2 \cite{sandler2019mobilenetv2}, RexNet100 \cite{han2020rexnet}, ResNet50 \& ResNet101 \cite{he2015deepresidual}, ResNest101e \cite{zhang2020resnest}, InceptionV3 \& InceptionV4 \cite{szegedy2014going}, and YOLOv8 \cite{Ultralytics2023}. Additionally, we incorporated recent FGVR architectures for a holistic performance assessment. Selections are based on publication date and benchmark performance on fine-grained datasets: SIM-Trans \cite{sun2022simtrans}, CMAL-Net \cite{LIU2023109550}, ViT-NeT \cite{kim2022vitnet}, IELT \cite{Xu2023}, TransFG \cite{he2021transfg}, MetaFormer \cite{diao2022metaformer}, CAP \cite{behera2021cap}, FFVT \cite{wang2022featurefusion}, and HERBS \cite{chou2023hightemperature}.

For training, all deep neural networks used a consistent input resolution of \(384\) pixels due to the Swin-T backbone in the plug-in module. The training involved a standardized regimen with a batch size of \(16\) over \(100\) epochs. The plug-in module initially employed the SGD optimizer (\(\eta = 5 \times 10^{-4}\)), later switching to the LION optimizer (\(\eta = 5 \times 10^{-6}\)). Other neural networks maintained a consistent learning rate (\(\eta = 5 \times 10^{-3}\)) with the AdamW optimizer.

\section{\textbf{Results \& Discussion}}
\subsection{Ablation Analysis}
We commence by presenting outcomes originating from the base Plug-in module, followed by the incorporation of the LION optimizer. Subsequently, we highlight findings from aiming to fine-tune network components for enhanced wrist pathology classification. This process yields three variants of the plug-in module, and we proceed to evaluate how their combination in a unified network through majority voting compares with both conventional and recent FGVR techniques. The accuracy of the base and post-integration LION network is shown in Table \ref{tab:table2} on both test sets. 

\begin{table}[h]
\centering
\caption{ Evaluation of plug-in module after integration of LION optimizer. }
    \begin{tabular}{l c c}
    \hline
    Model & Test Set 1 & Test Set 2\\
    \hline 
  
  PIM  & 84.38\% & 82.50\%\\
  PIM + LION & \textbf{85.44}\% & \textbf{83.75\%}\\
    \hline               
  \end{tabular}
  \label{tab:table2}
\end{table}

The results indicate that the accuracy of the plug-in module improves on both test sets when incorporating the LION optimizer. This improvement is credited to the enhanced generalization capacity of the plug-in module after integrating LION, resulting in superior performance on unseen data. The improved generalization is associated with the sign operation in LION, introducing noise to the updates, acting as a regularization technique, and contributing to better generalization.

We turn our attention to two critical aspects of the plug-in module: \enquote{Number of Selections} and \enquote{FPN Size}. The \enquote{Number of Selections} refers to the count of regions extracted from each backbone block by the Weakly Supervised Selector. On the other hand, \enquote{FPN Size} plays a vital role in determining the dimensions of the feature maps processed within the GCN. The value assigned to the projection size (derived from the FPN Size) extensively influences the organization of the network's layers. Table \ref{tab:table5} provides an evaluation of the plug-in module across different sets of selections. The default configuration demonstrates the highest accuracy, and consequently, we have chosen to retain the default selection of areas.

\begin{table}[h]
\centering
\caption{ Evaluation of plug-in module on different arrays of selections. }
    \begin{tabular}{l c c }
    \hline
    Number of Selections & Test Set 1 & Test Set 2 \\
    \hline 
    
  (256,128,64,32) & 81.70\% & 81.25\%\\
  (512,256,128,64) & 83.12\% & 82.50\%\\
  (1024,512,128,64) & 83.80\% & 81.25\%\\
  (1024,512,128,128) &  84.81\% & 81.25\%\\
  (2048,512,128,128) & 82.10\% & 78.75\%\\
  (2048,512,128,32) (default) & \textbf{85.44\%} & \textbf{83.75\%}\\
  (2048,512,128,64)  & 84.60\% & 82.50\%\\
    \hline               
  \end{tabular}
  \label{tab:table5}
\end{table}

Table \ref{tab:table6} presents the evaluation of the plug-in module across various FPN sizes selected arbitrarily. This analysis seeks to determine whether adjusting the FPN size, either increasing or decreasing, enhances the module's performance on this specific dataset. The results indicate that an FPN size of 1024 achieves the highest accuracy on the augmented test set, while the default FPN size achieves the highest accuracy on the original test set.

\begin{table}[h]
\centering
\caption{ Evaluation of plug-in module on different sizes of FPN. }
    \begin{tabular}{l c c}
    \hline
    FPN Size & Test Set 1 & Test Set 2 \\
    \hline 
    
  512 & 82.91\% & 78.75\%\\
  1024 & \textbf{85.70\%} & 81.25\%\\
  1536 (default) & 85.44\% & \textbf{83.75\%}\\
  2048 &  85.23\% & 81.25\%\\
  3000 & 81.01\% & 80.00\%\\
    \hline               
  \end{tabular}
  \label{tab:table6}
\end{table}

Table \ref{tab:table7} summarizes the incremental accuracy enhancements achieved with the plug-in module. A significant performance improvement is evident with the augmentation of training data. The subsequent integration of the LION optimizer further enhances performance across both sets. Ablation analysis indicates that an FPN size of 1024 improves the accuracy of the PIM on the augmented test set, albeit with a slight decrease in accuracy on the original test set. Lastly, our ensemble approach surpasses the performance of all three individual configurations.

\begin{table}[h]
\centering
\caption{ Improvements in Accuracy of Plug-in Module for FGVR. }
    \begin{tabular}{l c c}
    \hline
    Model Variant & Test Set 1 & Test Set 2\\
    \hline 
  PIM (Base)  & 84.38\% & 82.50\%\\
  PIM + LION & 85.44\% & 83.75\%\\
  PIM + LION + 1024 FPN & 85.70\% & 81.25\%\\
 \textbf{Model Ensemble (Our Approach)} & \textbf{87.34\%} & \textbf{83.75\%}\\
    \hline               
  \end{tabular}
  \label{tab:table7}
\end{table}

Table \ref{tab:table9} shows the evaluation of the plug-in module using different configurations alongside our ensemble approach using sensitivity, specificity, and precision on test set 2 (challenging set).

\begin{table}[h]
    \centering
    \caption{Sensitivity, specificity, and precision scores on test set 2.}
    \begin{tabular}{ccccc}
        \hline
        Config & Class & Sensitivity & Specificity & Precision\\
        \hline
        \multirow{4}{*}{PIM (Base)} & 0 & 53\% & 100\% & 100\%\\
        & 1 & 100\% & 78\% & 68\% \\
        & 2 & 93\% & 98\% & 93\%\\
        & 3 & 74\% & 96\% & 89\%\\
        \hline
        \multirow{4}{*}{PIM+LION} & 0 & 71\% & 98\% & 92\%\\
        & 1 & 96\% & 87\% & 77\% \\
        & 2 & 93\% & 98\% & 93\%\\
        & 3 & 74\% & 93\% & 81\%\\
        \hline
        \multirow{4}{*}{PIM+LION+1024FPN} & 0 & 65\% & 95\% & 79\%\\
        & 1 &  92\% & 87\% & 77\% \\
        & 2 & 93\% & 98\% & 93\%\\
        & 3 & 74\% & 93\% & 81\%\\
        \hline
        \multirow{4}{*}{Our Approach} & 0 & 75\% & 98\% & 91\%\\
        & 1 & 98\% & 95\% & 86\% \\
        & 2 & 90\% & 99\% & 96\%\\
        & 3 & 86\% & 92\% & 78\%\\
        \hline
    \end{tabular}
    \label{tab:table9}
\end{table}

\begin{figure*}
\centering 
\includegraphics[width=0.6\textwidth, height=3.5cm]{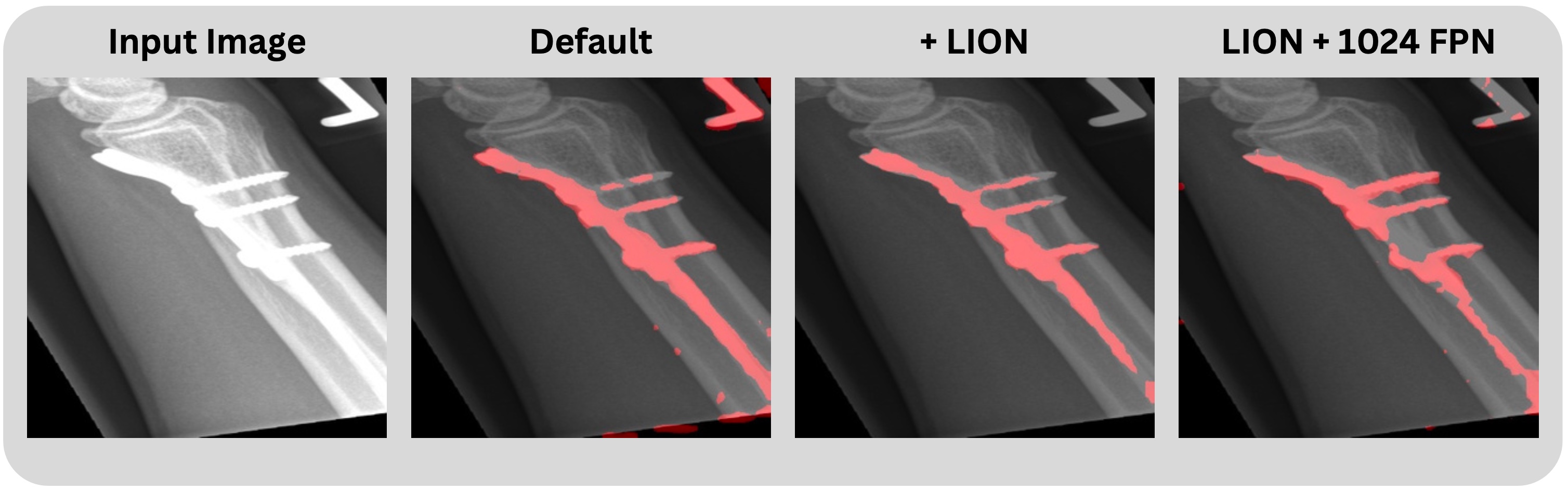}
\caption{Heatmaps produced from the three employed configurations of the plug-in module for FGVR.}
\label{fig1:heatmaps}
\end{figure*}

Fig. \ref{fig1:heatmaps} illustrates heatmaps generated by the three configurations of the plug-in module, trained on a limited dataset, with an aim to visualize the discriminative regions influencing the classifications. In the sample image representing the metal class, despite image over-exposure, the plug-in module successfully identified the class and emphasized the presence of metal. Notably, in many cases, conventional standards would render such an image undiagnosable by radiologists \cite{Atkinson2019}. The heatmap generated by default configuration shows that, while highlighting the metal, it also emphasizes non-discriminative regions, such as the letter \enquote{L} indicating the side of the arm. In contrast, the configuration after LION integration exclusively highlights the metal.

\subsection{Evaluation Against Other Deep Convolutional Neural Networks}
The evaluation results for each deep neural network, the base plug-in module, and our ensemble approach are presented in Table \ref{tab:table11}. The upper section of the table presents accuracies for conventional CNN techniques, while the lower section includes the performances for FGVR techniques. As hypothesized, the plug-in module for fine-grained recognition outperforms all state-of-the-art neural network architectures on the specific task of wrist pathology recognition. To further demonstrate the discriminative ability of our approach on this fine-grained task, we assess its performance using the original test set, termed the challenging test set, which consists of a mere 80 images without any augmentation. From Table \ref{tab:table11}, we select the top 5 high-performing models for this evaluation, and the results are shown in Table \ref{tab:table10}. It can be seen that our ensemble approach demonstrates superior performance even when the test set is highly limited. 

\begin{table}[h]
\centering
\caption{Performance evaluation of different deep neural networks along with our ensemble approach for FGVR on test set 1.}
    \begin{tabular}{l c c}
    \hline
    Model & Test Accuracy (Test set 1)\\
    \hline 
  EfficientNetV2 & 53.59\% \\
  VGG16 & 65.82\% \\
  ViT & 70.25\% \\
  RegNet & 72.36\% \\
  DenseNet201 & 73.42\% \\
  MobileNetV2 & 76.37\% \\
  ResNet101 & 77.43\% \\
  DenseNet121 & 78.21\% \\
  ResNest101e & 78.27\% \\
  InceptionV4 & 78.69\% \\
  ResNet50 & 79.11\% \\
  InceptionV3 & 79.54\% \\
  EfficientNet\_b0 & 79.96\% \\
  YOLOv8x & 80.50\% \\
  \hline
  SIM-Trans & 72.79\% \\
  CMAL-Net & 76.58\% \\
  ViT-NeT & 76.79\% \\
  IELT & 78.10\% \\
  TransFG & 78.90\% \\
  MetaFormer & 78.90\% \\
  CAP & 80.80\% \\
  FFVT & 81.65\% \\
  HERBS & 82.70\% \\
  PIM (Base) & 84.38\% \\
  \textbf{Our Approach} & \textbf{87.34\%} \\
    \hline                                                      
  \end{tabular}
  \label{tab:table11}
\end{table}

\begin{table}[h]
\centering
\caption{Performance evaluation on the original unaltered test set.}
    \begin{tabular}{l c c}
    \hline
    Model & Test Accuracy (Test set 2)\\
    \hline 

      FFVT & 33.75\% \\
      CAP & 71.25\% \\
  YOLOv8x & 72.50\% \\
  
  HERBS & 78.75\% \\
  PIM (Base) & 82.50\% \\
  \textbf{Our Approach} & \textbf{83.75\%} \\
    \hline                                            
  \end{tabular}
  \label{tab:table10}
\end{table}

\section{\textbf{Conclusion \& Future Work}}
In this study, we have showcased the effectiveness of addressing wrist pathology classification as a fine-grained problem on a highly limited dataset. We identified an optimal architecture for this task and introduced two variations to form an ensemble approach that surpassed all other state-of-the-art methods used in the study. Looking forward, one direction involves refining fine-grained architectures specifically tailored for wrist pathology recognition. Although existing networks tackle general fine-grained issues, we anticipate that future developments in our approach could potentially eliminate the need for manual annotation. Despite training our network on a limited dataset, we observed high-quality heatmaps. We suggest that employing larger datasets with only image-level annotations could further improve heatmap quality.

\section{Acknowledgement}
This work was supported by the Curricula Development and Capacity Building in Applied Computer Science for Pakistani Higher Education Institutions (CONNECT) Project NORPART-2021/10502, funded by the Norwegian Directorate for Higher Education and Skills (DIKU).

\bibliographystyle{IEEEtran}
\bibliography{cas-refs}

\color{red}

\end{document}